\pdfoutput=1

\documentclass[11pt]{article}

\usepackage[dvipsnames]{xcolor}

\usepackage{acl}
\usepackage{tabularx}

\usepackage{times}
\usepackage{latexsym}
\usepackage{booktabs}
\usepackage{multirow}
\usepackage{multicol}
\usepackage{graphicx}
\usepackage{subcaption}
\usepackage{hyperref}
\usepackage{tabularx}
\usepackage{tikz}

\usetikzlibrary{shapes,arrows,positioning, fit}

\usepackage[T1]{fontenc}

\usepackage[utf8]{inputenc}

\usepackage{microtype}
\usepackage{linguex}

\title{Hybrid Human-LLM Corpus Construction and LLM Evaluation for Rare Linguistic Phenomena}

\author{Leonie Weissweiler, Abdullatif Köksal, Hinrich Schütze\\
  LMU Munich \& Munich Center for Machine Learning\\
  \texttt{weissweiler@cislmu.de} \\}

\newcounter{notecounter}
\newcommand{\enotesoff}{\long\gdef\enote##1##2{}}
\newcommand{\enoteson}{\long\gdef\enote##1##2{{
			\stepcounter{notecounter}
			{\large\textbf{ \hspace{1cm}\arabic{notecounter} $<<<$ ##1: ##2 $>>>$\hspace{1cm}}}}}}
\enoteson
\enotesoff

\def\dnrm#1{\mbox{$_{\hbox{\scriptsize #1}}$}}

\def\eqref#1{Eq.~\ref{eqn:#1}}
\def\eqlabel#1{\label{eqn:#1}}

\begin{document}
\maketitle
\begin{abstract}
Argument Structure Constructions (ASCs) are one of the most
well-studied construction groups, providing a unique
opportunity to demonstrate the usefulness of Construction
Grammar (CxG).  For example, the caused-motion construction
(CMC, ``She sneezed the foam off her cappuccino'')
demonstrates that constructions must carry meaning,
otherwise the fact that ``sneeze'' in this context causes movement cannot be explained.  We form the hypothesis that this
remains challenging even for state-of-the-art Large Language
Models (LLMs), for which we devise a test based on
substituting the verb with a prototypical motion verb.  To
be able to perform this test at statistically significant
scale, in the
absence of adequate CxG
corpora, we develop a novel pipeline of NLP-assisted 
collection of linguistically annotated text.  We show how dependency parsing and GPT-3.5 can be used to
significantly reduce annotation cost and thus enable the
annotation of rare phenomena at scale. We then evaluate GPT, Gemini,
Llama2 and Mistral models for their understanding of the CMC
using the newly collected corpus. We find that all models
struggle with understanding the motion component that the CMC adds to a sentence.
\end{abstract}

\section{Introduction}

\ex.\label{cm_1} She sneezed the foam off her cappuccino.

\ex.\label{cm_2} They laughed him off the stage.

These are two examples of
the caused-motion
construction (CMC)
in which the verb
behaves unusually: \textit{sneeze}
and \textit{laugh} typically do not take multiple arguments,
nor do they typically convey that something was moved by sneezing/laughing. This poses a challenge
to
any naive form of lexical semantics:
it would not make sense for someone writing a
dictionary to include, for each
intransitive verb, the meaning and valency of the CMC.
Almost any verb can appear in the CMC
as long as we
can imagine a scenario in which the action it describes causes motion. The fact
that humans easily understand the CMC showcases a
main feature of Construction
Grammar \citep{croft2001radical, goldberg1995}: the meaning
is attached to the construction itself, and not the verb.
Putting the verb into this construction adds the new meaning and valency. This is
one reason that constructions pose a   challenge  to Large
Language Models (LLMs), as they would have to learn to
attach the meaning to this construction and retrieve it when
necessary. Its extreme rareness and productivity
makes it impossible to memorise all instances
and memorization would not be sufficient because the meaning shift to the verb is creative and is
influenced by the specific context.

The research questions of this paper
therefore are: Have LLMs learned the meaning of the CMC and
how can we construct the resources needed to determine the
status of CMC in LLMs?

We first address the second question, of collecting data for this at scale. This is challenging for several reasons. First, the
CMC is a very rare phenomenon. Second, we are mostly
interested in instances that are non-prototypical,
i.e., where the verb does not typically encode motion, unlike
e.g. `kick' or `throw'. 
Third, this construction cannot be automatically identified
using only syntactic criteria: words might be in the correct syntactic slots required by the CMC,
but not create a CMC reading if the semantics of the sentence do not fit. For example, ``I would take that into account'' is structurally identical to the examples above, but nothing is moving.

This shows that there is a crucial semantic
component.
The rarity makes it infeasible to manually sift through a corpus to collect
a dataset of the CMC, while the semantic complexity makes it infeasible
to do so fully automatically.

In this way, we consider the CMC exemplary of rare phenomena
of language that have been largely set aside in
Computational Linguistics and in recent evaluation of LLMs
in particular. This may be due to them being considered
the \textit{periphery} of language, rather than the
core \cite{chomsky1993lectures}, or simply due to the
described difficulty in finding appropriate data to
investigate both the phenomena and their representation in
LLMs. However, it is our point of view that as the
performance of such models increases across the board, it is
vital to turn to ``edge cases'' to accurately identify
performance gaps. This is particularly important as  rare
phenomena
may be indicators of  systematic underlying
problems of an NLP paradigm.

To study rare phenomena, we need data.
To this end, in section \ref{pipeline_section}
we propose a novel
annotation pipeline that combines
dependency parsing with
an advanced LLM,
GPT-3.5 \citep{openai2023chatgpt}. 
The aim of our pipeline is to minimise the cost of running GPT-3.5 and compensating human annotators, while maximising the number of positive, manually verified, linguistically diverse instances in the dataset.

We further use
insights about the semantic regularities in the CMC to
automatically expand this to a second corpus of 127,955
sentences, which have not been manually verified, but
with high likelihood are also CMC instances.
After creating our corpus, we now return to our aim of evaluating state-of-the-art LLMs for their understanding of the CMC, as an example of a semantically challenging ``edge case''.

In Section \ref{sec:eval}, we therefore develop a test for different LLMs' understanding of the CMC, by
giving an instance and asking if
the direct object is physically moving. 
We then replace the
non-prototypical verb (e.g., ``sneeze'') by a prototypical one that always
encodes motion (e.g., ``throw'') and ask the model again if the direct object is moving. We expect models that do not fully understand the CMC to fail to consistently answer both questions with ``yes''.
We observe that
all models struggle with this task. Mixtral 8x7b performs best, but still has an error rate of over 30\%.

We make three main contributions:
\begin{itemize}
    \item
We propose a hybrid human-LLM corpus construction method
and show its effectiveness for non-prototypical CMC, an
extremely rare phenomenon.
We discuss how our design and our guidelines can be
applied to data collection needs for other
linguistic phenomena.
\item We release a corpus of manually verified instances of the CMC of 765 sentences and a second, semi-automatically annotated corpus of 127,955 sentences.
    \item We evaluate different sizes of GPT, Llama2, Mistral, and Gemini models on their understanding of the CMC and find that all models perform poorly, with the best model only achieving 69.75\%.
\end{itemize}

\section{Related Work}

\textbf{Evaluation of LLMs' Understanding of Constructions.}
\citet{tayyar-madabushi-etal-2020-cxgbert}
conclude that
BERT \cite{devlin-etal-2019-bert} can
classify whether two sentences contain instances of the same
construction.
\citet{tseng-etal-2022-cxlm}
show that LMs have higher prediction accuracy on fixed 
than on variable syntactic slots 
and infer that
LMs acquire  constructional
knowledge (i.e., they understand the ``syntactic context''
needed to identify a fixed slot).
\citet{weissweiler-etal-2022-better} find that
LLMs  reliably discriminate instances of the
English Comparative Correlative (CC) from 
superficially similar contexts.  However, LLMs do not
produce correct inferences from them, i.e., they do not
understand its meaning. 
 
Most related to this work, \citet{li-etal-2022-neural} probe
for LMs' handling of four ASCs: ditransitive, resultative,
caused-motion, and removal.
They
adapt the findings of \citet{bencini2000}, who used a
sentence sorting task to determine whether human
participants perceive the argument structure or the verb as
the main factor in the sentence meaning. 
They find
that, while human participants prefer sorting by the
construction more if they are more proficient English
speakers, PLMs show the same effect in relation to training
data size.  
In a second experiment, they then insert random
verbs that are incompatible with one of the constructions,
and measure the Euclidean distance between the verbs'
contextual embedding and that of a verb that is prototypical
for the construction. 
They demonstrate that construction information
is picked up by the model, as the contextual embedding of the verb
is brought closer to the corresponding prototypical verb embedding.

\citet{mahowald2023}
investigates GPT-3's \cite{brown2020language} understanding
of the English AANN
construction, assessing its grasp of the construction's
semantic and syntactic constraints. Utilising a few-shot
prompt based on the CoLA corpus of linguistic
acceptability \citep{warstadt-etal-2019-neural}, he creates
artificial AANN variants as probing data. GPT-3's
performance on the linguistic acceptability task is found to align with human judgments across most conditions.

\textbf{Linguistic Annotation with GPT.}
Since the release of ChatGPT, numerous papers have proposed
to use it as an annotator.
\citet{gilardi2023chatgpt} find that ChatGPT outperforms crowd-workers on tasks such as topic detection. \citet{yu2023assessing} and \citet{savelka2023unreasonable} evaluate the accuracy of GPT-3.5 and GPT-4 against human annotators, while \citet{koptyra2023clarin} annotate a corpus of data labelled for emotion by ChatGPT, but acknowledge its lower accuracy compared to a human-annotated version. In the area of Construction Grammar, \citet{torrent2023copilots} use  ChatGPT to generate novel instances of constructions.

Most related to our work are papers that propose a cooperation between the LLM and the human annotator. \citet{holter2023human} create a small gold standard for industry requirements by generating an initial parse tree with GPT-3 and then correcting it with a human annotator. \citet{pangakis2023automated} investigate LLM annotation performance on 27 different tasks in two steps. First, annotators compile a codebook of annotation guidelines, which is then given to the LLM as help for annotation, and then the codebook is refined by the annotators in a second step. However, they find little to no improvement from the second step. \citet{gray2023can} make an LLM pre-generate labels for legal text analytics tasks which are then corrected by human annotators, but find that this does not speed up the annotation process. 

In contrast, our work proposes a hybrid human-LLM pipeline
that minimizes
the cost of dataset creation. We emphasise
prompt design and engineering, a critical factor  in
effective use of LLMs.

\textbf{Computational Approaches to Argument Structure Constructions (ASCs).}
In addition to  probing work discussed above, ASCs have
also been studied from a computational perspective.
\citet{kyle-sung-2023-argument} 
leverage a UD-parsed corpus as
well as FrameNet \cite{ruppenhofer2016framenet} semantic
labelling to annotate a range of ASCs. 

\citet{hwang-palmer-2015-identification} 
identify CMCs and four different subtypes based on
 linguistic features. Some of these are automatically
 generated, but others are gold annotations. This limits the applicability to large, unannotated corpora. 

\citet{hwang2023automatic} conduct an automatic analysis of constructional diversity to predict ESL speakers' language proficiency. Similar to our first filtering step, they perform an automatic dependency parse and then identify a range of constructions, including the CMC, using a decision tree built on the parse. They do not employ any further filtering.

\begin{figure*}
    \centering
    \includegraphics[width=\linewidth]{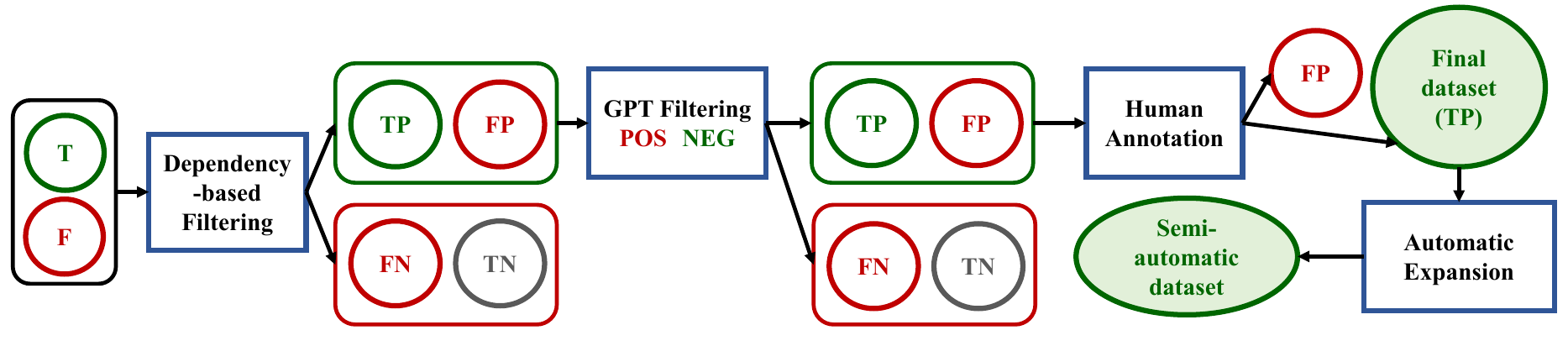}
    \caption{Flowchart of our annotation pipeline. For details of each step refer to \S\ref{pipeline_section}.}
    \label{fig:pipeline}
\end{figure*}

\section{Data Collection}
\label{pipeline_section}
We address the following problem setting.

A \textbf{linguistic researcher} (a computational or corpus
linguist) wants to investigate a \textbf{rare phenomenon}.
They want
to find as many examples of the phenomenon as possible, which must be
verified by human annotators. There are three tools at their
disposal: linguistic resources, GPT-3.5, and human
annotators. The human annotators will generally be experts
in linguistic annotation.

Our key idea is that data collection will proceed in a pipeline, where a corpus
is first filtered using dependency parsing and the syntactic constraints of the phenomenon,
the output is further filtered with prompt-based classification using GPT, and the sentences
which it labels as positive are then manually annotated by a human. Each step in the pipeline is meant to further concentrate the rate of instances in the corpus that will then be manually annotated, therefore reducing total annotation effort.

The main cost of data collection is the cost for the GPT-3.5
API and for human annotators. 
We assume that any
expenses for linguistic resources and the computational
infrastructure at our disposal are negligible in comparison.
\emph{Our aim is to minimise the cost for GPT-3.5 and annotators
while maximising the number of positive, manually verified, diverse
instances.} We assume that the use case of the linguistic resource is such that it needs to be manually annotated, so that we cannot simply fully rely on an LLM.

We propose a way of computing the cost for this problem setting and
a pipeline for producing a novel linguistic resource while minimising cost.

Our main goal is minimising the cost per confirmed
instance; however, we also have a secondary goal:
the final set of instances should be diverse.
Regardless of the specific goals
of the linguistic researcher, it is unlikely that they would
be served by a set of sentences that do not represent the
true diversity of the  phenomenon. 
Extreme cost-minimising
measures -- such as making the dependency filtering rules
described in \S\ref{ud_section} too strict or asking
GPT-3.5 to provide examples of the phenomenon -- would
therefore be counterproductive.

The baseline here is to take an annotator,
give them a corpus, set them on the task of reading
through it and marking all sentences that contain instances
of the phenomenon.
Let $C\dnrm{HR}$ be the
fixed cost of the
human review of one sentence and $N$ the size of the corpus.
Then the total cost of creating the resource is
$J\dnrm{naive} = C\dnrm{HR} N$ as the annotator
has to sift through all sentences. 
Our problem setting is that
the phenomenon is rare, i.e., the corpus contains very few true
positives. Thus, $\mbox{TP} \ll N$
and the number of positive instances
$\mbox{TP}$
found after annotating the corpus will be small.
This makes this method cost-prohibitive. We therefore turn to dependency parsing
with spacy \cite{spacy2} for prefiltering.

\subsection{Step 1: Dependency Parsing}
\label{ud_section}

\subsubsection{General Considerations}

Figure \ref{fig:pipeline} shows our pipeline.  In the first
step, we dependency-parse the corpus and apply a
pattern to filter out all sentences that, with high
likelihood, are not instances of the phenomenon.

For this dependency annotation, we could
rely on annotated treebanks such as Universal Dependencies \cite{demarneffeUD}. But to find a diverse and sufficiently large set of instances, particularly in languages other than English, available treebanks may not be large enough for the rare phenomenon that we are targeting. 

We therefore turn to automated dependency parsing to annotate large amounts of data. We assume that this part of the pipeline has negligible
cost, as it is carried out either through a web service or on
existing infrastructure.

After dependency parsing, we want filters that
preserve the diversity of the found sentences. We therefore design subtree filters that
preserve recall above all else. This is especially advisable
as parsing will lead to some parsing errors that we want to
be tolerant of, and the rarer the phenomena that we are
targeting, the more likely they are to be parsed
incorrectly, as even human annotation guidelines might be
inconsistent for them.

To write the patterns, we recommend to start with a list of
gold instances, dependency parse them and manually look at the similarities. If one wants to work only with a
treebank, it might be advisable to start at first with a
highly lexicalised version of the phenomenon to find a first
example sentence and go on from there. In terms of tools, if
one does not want to write the code for finding the
instances themselves, available tools include
GrewMatch,\footnote{\url{https://match.grew.fr/}}
DepEdit \cite{peng-zeldes-2018-roads}, Corpus
Workbench\footnote{\url{https://cwb.sourceforge.io/index.php}}
and SPIKE \cite{shlain-etal-2020-syntactic}.

\begin{figure}
    \centering
    \includegraphics[width=\linewidth]{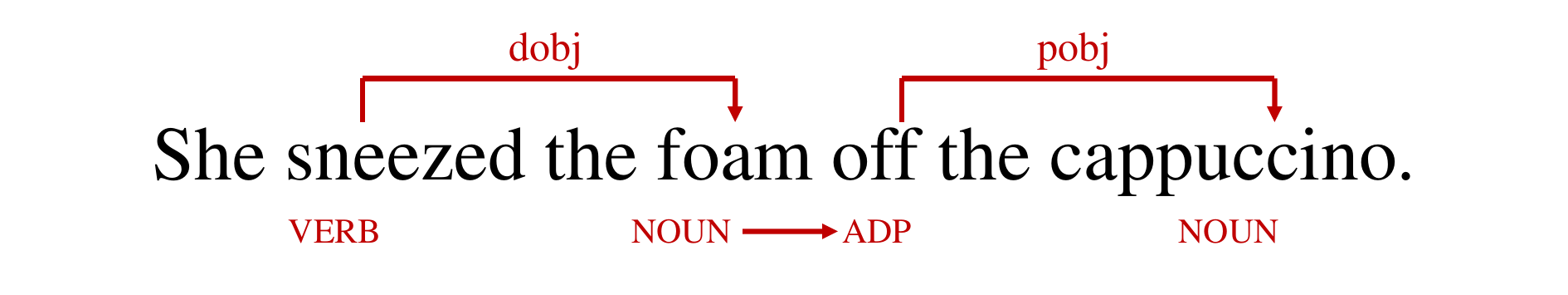}
    \caption{Our dependency subtree
for the Caused Motion Construction (highlighted in red),
    shown with an example.
    The straight arrow (NOUN $\rightarrow$ ADP) indicates
    that the words must be adjacent.}
    \label{fig:ud_tree}
\end{figure}

\subsubsection{Designing a Dependency Subtree for the CMC}
Figure \ref{fig:ud_tree} shows the dependency subtree we use
as a filter for
CMC.
It was designed by
automatically parsing examples of positive and negative
instances from \citet{goldberg1992}, and manually designing
subtrees that cover the positive instances except for
clear parsing errors, while excluding as many of the
negative examples as possible. We run an automatic
dependency parser from the spacy toolkit\footnote{version 3.2.0} on
the reddit corpus \cite{reddit} and extract all matching sentences. This also allows us to
extract the location of the potential CMC instance and its
parts as a side product of the filtering step: We extract
the sentence, the lemmatised verb, direct object,
preposition, and prepositional object, as well as their
positions in the sentence.

\subsection{Step 2: Prompt-based Few-shot Classification with GPT}
\label{cost_section}

\subsubsection{General Considerations}
Even after dependency-based filtering, the positive instances would still be
very rare in the output and it is therefore not feasible that the output 
is directly annotated by a human. We therefore introduce a further filtering
step with GPT to ``concentrate'' the positive instances even more (Step 2 of the pipeline, Figure \ref{fig:pipeline}), i.e. we want GPT to remove most
negative instances while keeping as many
positive instances as possible. The remaining data can then
be cost-effectively annotated by the human annotator.
The aim is to reduce the cost per instance (i.e., cost per true positive, TP) 
as much as possible.

There are two components of the cost: the cost of querying
GPT and the cost of human annotation.
Our two key ideas are:
\begin{itemize}
\item We consider the two costs jointly and optimize the
pipeline for overall lowest cost per TP.
\item Design and selection of the prompt used with the API
is a major determinant for the cost of the pipeline. We
propose a workflow for creating effective prompts.
\end{itemize}

A particular prompt may require many tokens in total,
thereby incurring a higher API cost. But it may also have
high accuracy, thereby reducing the cost of human
annotation. We jointly consider both cost components when
designing and selecting prompts.

\paragraph{Minimizing the cost per true positive}
Given an annotated development set $V$ that is representative of
the corpus that we want to annotate with our pipeline, let
$J(C\dnrm{HR},C\dnrm{API},i)$ be the cost per true positive
where
$C\dnrm{HR}$ is the human annotation cost
per sentence,
$C\dnrm{API}$ is the cost of processing an input/output token with the
API and $i$ (for instruction) is a prompt.
We can then estimate
$J(C\dnrm{HR},C\dnrm{API},i),$ the cost per true positive,
as follows:
\begin{equation}
\frac{C\dnrm{API}t(V,i)+C\dnrm{HR}(\mbox{TP}(V,i)+\mbox{FP}(V,i))}
{\mbox{TP}(V,i)} \eqlabel{costpertp}
\end{equation}
where
we process the
development set using the API and prompt $i$
and record: $\mbox{TP}(V,i)$,
the number of true positives,
$\mbox{FP}(V,i))$,
the number of false positives,
and
$t(V,i)$,
the sum of the
number of tokens input to the API and the number of tokens
returned by the API.

We create a variety of different prompts $i$ and then select
our final prompt $i'$
as  the one with the lowest per-TP cost:
\[
i' = \mbox{argmin}_i J(C\dnrm{HR},C\dnrm{API},i)
\]

\paragraph{Determining the size of the input corpus}
When applying our pipeline, we often will have a requirement
for a specific number $\mbox{TP}\dnrm{req}$ of the
phenomenon, e.g.,
$\mbox{TP}\dnrm{req} = 1000$
instances of the CMC.
After selecting a prompt $i$ and determining
$\mbox{TP}(V,i)$ on the development set, we can estimate 
the size $N$ of the input corpus that will result in a set
of
$\mbox{TP}\dnrm{req}$ instances to be
output by the pipeline as:
\[
N := |V|\frac{\mbox{TP}\dnrm{req}}{\mbox{TP}(V,i)}
\]

\subsubsection{Prompt Engineering and Few-shot Selection}
\label{prompt_suggestions}

We suggest to first manually annotate a development set of
positive and negative instances.
The development set should be as representative of the
corpus to be annotated as possible, including the rate of $\mbox{TP}$s.
This development set can then
be used, along with a small fixed budget, to develop prompts
and finally select the 
best-performing prompt.
We
present possible strategies for prompt development, using CMC
as the running example.

\paragraph{Development Set}
For creating the development set $V$, we manually annotate 504 (133 $P$, 371 $N$) sentences from
the output of the dependency filtering step. To ensure that
$V$ is both diverse and relevant, we group
the prefiltered dataset by verb, and starting with the
highest-frequency verbs, take at most 5 positive and 5
negative sentences from every verb, where no preposition
appears twice in either the positive or the negative
sentences selected. We
choose 5 shots from each class to be included as examples in
the prompt, which are a mixture of sentences taken
from \citet{goldberg1992} and additional data that was not
used for $V$.  Using this development set $V$, 
we can determine TP$(V,i)$, FP$(V,i)$ and $t(V,i)$
of each prompt and compute its per-TP cost using \eqref{costpertp}.

\paragraph{Iterative Prompt Development}

We start with a simple base prompt and iteratively attempt improvements to it. The total cost of this experimentation was about \$10. The full details of all attempted prompts are given in the appendix in Section \ref{prompt_details}.\footnote{The base model used was gpt-3.5-turbo-1106, the GPT-4 model was gpt-4-0125-preview.}

Most improvements on the base prompt are uncontroversial, as
they decrease the total cost under the assumption that
$C\dnrm{HR}$ is not less than \$0.0001 (and the
cost per API token is dominated by the cost of human annotation). These include:
\begin{itemize}
\setlength\itemsep{0.2em}
    \item long task description
    \item long system prompt
    \item JSON format for input and output
    \item few-shots alternate by class
    \item explanations for the labels added to the few-shots and demanded from the model
    \item only 10 sentences classified at one time
\end{itemize}

\begin{table*}
    \centering
    \small
    \resizebox{\textwidth}{!}{
    \begin{tabular}{rlrrrrrrrrrr}
    \toprule
         &  & &  & \multicolumn{2}{c}{\textbf{Sent's to
    Annotate}}&  & \multicolumn{4}{c}{\textbf{Total Cost}}\\ 
         
        \cmidrule(lr){5-6} \cmidrule(lr){8-11}
          \textbf{P}&  \textbf{Details} & \textbf{Prec.}
          & \textbf{Rec.} & \textbf{LLM} & \textbf{Human}&
          \textbf{API} 
          & $C\dnrm{HR}$=\textbf{\$.001} & $C\dnrm{HR}$=\textbf{\$0.2 } & $C\dnrm{HR}$=\textbf{\$1} & $C\dnrm{HR}$=\textbf{\$2}\\
        \cmidrule(lr){1-4} \cmidrule(lr){5-6} \cmidrule(lr){7-7} \cmidrule(lr){8-11}
         5& Simple&56.54 & 71.42 &  5,304  & 1,768 & \$0.4 & \textbf{\$2.2}  & \$177 & \$1,768 & \$3,537\\
         12& 5 + expl. &83.75 & 50.37 &  7,522 & 1,194 & \$1.4 & \$2.6  & \textbf{\$120} &\$1,195 & \$2,389\\
         17& 12 + GPT-4 &90.09 & 81.96 & 5,040 & 1,110& \$16.2 & \$17.3 & \$127 & \textbf{\$1,126}  &\$2,236\\
          18& 17 + best-of-3&91.81 & 75.93 & 4,989  & 1,089 & \$48.3 & \$49.4  & \$157  & \$1,137 &\textbf{\$2,226}\\
        \cmidrule(lr){1-4} \cmidrule(lr){5-6} \cmidrule(lr){7-7} \cmidrule(lr){8-11}
          - & Human only & - & - & - & 3,789 & \$0.0 & \$3.8 & \$575 & \$3,789 & \$7,578\\
    \bottomrule
    \end{tabular}
    }
    \caption{A comparison of the four best-performing
          prompts (5, 12, 17, 18) for different values of $C\dnrm{HR}$.
\textbf{P} = Prompt.
We give numbers
(sentences that need to be annotated by LLM/human)
for a scenario in which the desired size of
the final resource (output of pipeline when applied to the
raw corpus) is $N=1000$.
The human baseline depends solely on the rate of TPs (which is
higher here
than for the raw corpus to be processed by the pipeline
as the development set contains more positive instances).}
    \label{tab:final_prompts}
\end{table*}

Table \ref{tab:final_prompts}
illustrates the prompt development process and how the
choice of most cost-effective prompt depends on human cost
$C\dnrm{HR}$
and
API cost $C\dnrm{API}$. Prompts are numbered
(reflecting the sequence in which we created them).
Prompt 5 is a simple short prompt (consequently: low API
cost). Its precision is low and so the human has to
annotate more sentences to produce the final dataset of
$N=1000$ TPs of CMC.
Prompts 12, 17, 18 are consecutive refinements of
5. They result in more input to and output from the API 
because explanations are given with
the few-shots,
the model is asked to provide explanations and (for 18) we
classify each sentence three times (tripling the API
expense).
For 17 and 18, we use GPT-4, which is more capable, but also
more expensive than GPT-3.5.
Each refinement
improves precision, which then decreases
human cost.

The right four columns of the table illustrate how prompt
selection depends on human cost. For (unrealistic) very low
human cost ($C\dnrm{HR}$=\textbf{\$.001}), precision of the prompt does not matter much and
the simplest prompt (which has low API cost) is most
cost-effective. As human cost rises, the more expensive
prompts become competitive since now human cost is the
main factor and the API cost component becomes small in
relative terms. 
For $C\dnrm{HR}$=\textbf{\$.2}, prompt 12 is best;
for $C\dnrm{HR}$=\textbf{\$1}, prompt 17; and for 
$C\dnrm{HR}$=\textbf{\$2}, prompt 18.

As our \textbf{final prompt}, we select prompt 12 as it is
a good tradeoff between API cost and human cost.

\subsection{Final Dataset Collection}

\begin{table*}
    \centering
    \begin{tabular}{l}
    \toprule
        I \textcolor{ForestGreen}{laughed} \textcolor{Purple}{myself} \textcolor{NavyBlue}{onto} \textcolor{Mahogany}{the floor} .\\
        She had a bit of sauce on her lip so he \textcolor{ForestGreen}{kissed} \textcolor{Purple}{it} \textcolor{NavyBlue}{off of} \textcolor{Mahogany}{her} .	\\
        She was in pretty bad shape from drinking , so I \textcolor{ForestGreen}{helped} \textcolor{Purple}{her} \textcolor{NavyBlue}{into} \textcolor{Mahogany}{bed} and started the water .\\
        He \textcolor{ForestGreen}{rushed} \textcolor{Purple}{me} \textcolor{NavyBlue}{across} \textcolor{Mahogany}{the street} , and down the sidewalk west towards Bedlam Mental Hospital .\\
        I can \textcolor{ForestGreen}{pop} \textcolor{Purple}{my shoulder} \textcolor{NavyBlue}{ out of} \textcolor{Mahogany}{my socket} .\\
    \bottomrule
    \end{tabular}
    \caption{Examples from the final dataset. Verbs are highlighted in \textcolor{ForestGreen}{green}, direct objects in \textcolor{Purple}{purple}, prepositions in \textcolor{NavyBlue}{blue}, and prepositional objects in \textcolor{Mahogany}{red}.}
    \label{tab:dataset_examples}
\end{table*}
      
Using prompt 12, we collect a final dataset with a \$6
budget for the GPT-3.5-API and manually annotate the
output. Out of 20,408 sentences classified, 1303 are judged
by the model as positive. We annotate these by hand and find
632 positive instances. We combine these with the 133
positive instances from our development set for a final
number of 765 hand-annotated CMC instances.\footnote{We
hope to be able to release larger datasets using this method
in the future, given a larger computational and annotation
budget.}

We observe that  the 4-tuple of $<$verb, direct
object, preposition, prepositional object$>$ almost always
perfectly determines the class. We use this observation to
extrapolate from our manually annotated sentences to all
other sentences with the same 4-tuples and release a second
dataset of 127,955 high-likelihood CMC instances.

\section{Evaluation of LLMs' Understanding of the CMC}
\label{sec:eval}

\subsection{Methods}

The goal of our evaluation is to assess different LLMs for
their understanding of the CMC. The fact that the prompt
engineering above, even with 10 few-shot examples and
extensive task descriptions and explanations, is still far
from perfect, suggests that this is a challenging task
even for advanced models like GPT-3.5, which leads us to question if the same holds
for other LLMs.

Our LLM evaluation setup in this section differs from
prompt evaluation as we  do not explicitly refer to the
``caused-motion construction'', but rather prompt implicitly
for the model's understanding of the situation described. The key idea
is that in a CMC sentence, something is always physically
moving, even if the verb (e.g., ``sneeze'') does not
indicate this.
The distinction between 
prototypical vs.\ non-prototypical
instances is crucial here: for prototypical CMC instances
(``throw'', ``kick''),
the verb
already conveys the
meaning component of motion while for non-prototypical
CMC instances (``sneeze'', ``laugh'') it does not
and the LLM has to infer the additional meaning component of motion
from the construction.

Our setup is to ask ``In
the sentence \textit{sentence}, is \textit{direct\_object}
moving, yes or no?''. We then
replace the verb
of the CMC with the appropriately inflected form of ``throw'', and ask
the same question again, using the structural information extracted by the dependency filtering step.  We expect that models with no
understanding of the CMC would answer ``yes'' both times
only for prototypical instances, and switch from ``no'' to
``yes'' for non-prototypical ones. Models with a perfect
understanding of the CMC would always answer ``yes''. However, we do not group verbs into prototypical and
non-prototypical ones a priori, as we support the view that
this is on a gradient, and therefore the classes are not
clear-cut: verbs like ``brush'' or ``apply'' may not always
convey motion outside of the CMC, but certainly more often
than ``laugh''.

We conduct this experiment on our corpus of 765 hand-annotated sentences.
As API-based LLMs, we investigate GPT-3.5, GPT-4
\cite{openai2023chatgpt}, and Gemini Pro
\cite{geminiteam2023gemini}. From the family of open LLMs,
we further choose Llama2 \cite{touvron2023llama} with sizes
7b, 13b, and the quantised version of 70b, and Mistral 7b
\cite{jiang2023mistral} and Mixtral 8x7b
\cite{jiang2024mixtral}, as well as their respective
instruction-tuned versions. Models generate a sentence in response, which we then parse for versions of ``yes'' and ``no''.

\subsection{Quantitative Results}
Table
\ref{tab:gpt_results} presents the results in three
groups. (i) Y$\rightarrow$Y. The model
answers ``yes'' both times and therefore demonstrates that
it understands the CMC. (ii) N$\rightarrow$Y.
The model answers with ``no'' for the original sentence but
changes its answer to ``yes'' when the verb is changed to
``throw'', meaning that it does not understand the CMC.
(iii) X$\rightarrow$N. Even
with ``throw'', the model does not answer correctly that the
direct object is moving. We consider these to be general
failures of the model to understand the instruction, rather
than the CMC specifically.

The main result of this evaluation is that all models
perform poorly, with Mixtral 8x7b instruction-tuned (IT) performing best with
69.75\% and an overall error rate of 30.25\%
(12.13+18.12). Mistral 7b IT is a close
second. Surprisingly, GPT-4 follows at a distance of more
than 10 points at 57.07\%.  All other models perform worse.
This demonstrates that, despite the impressive progress LLMs
have recently made, they are still far from perfect natural
language understanding. Presumably, one reason is that
non-prototypical CMC instances (e.g., ``sneeze'') do not
occur frequently in the training data. In addition, this
type of construction is creative and complex use of language
that is harder to generalize to than the linguistic
behaviors that are tested in standard NLP benchmarks. Our
finding may be useful as guidance for the further
development of large language models. In addition to
semantic/pragmatic shortcomings of LLMs (e.g., on reasoning)
and ``bad behavior'' (output that is untruthful, offensive
etc.), it suggests that syntax and in particular the
syntax-semantics interface is also an area in which more
progress is needed for LLMs to come closer to human levels
of performance.

Surprisingly, our expectation that instruction-tuned models
will generally be better at understanding the premise of the
question (and therefore reduce
X$\rightarrow$N answers) only holds  for Mistral, but not for Llama2. Higher rates of X$\rightarrow$N
are generally associated with lower rates of
N$\rightarrow$Y, suggesting that models that understand the
question better also understand the CMC better. Most
surprisingly, the worst-performing models were the two GPT
models, Gemini Pro, and Mixtral 8x7b with instruction
tuning, while the best-performing models were smaller sizes
of Llama2.

\subsection{Qualitative Results}

We hypothesised that prototypical motion verbs would lead
to the model answering yes both times
(Y$\rightarrow$Y), as the original
sentence already obviously contains motion. It
might also make sense for these verbs to be more frequent in
the category of the model answering ``no'' both times (X$\rightarrow$N). While
this would not be the correct answer, a model that is
confused and answers incorrectly about a sentence with
``kick'' would likely give the same answer for ``throw''. We also
anticipated that the frequency of a model
changing its answer would be the higher,
the less prototypical the original verb is
(i.e., a ``no'' for highly non-prototypical ``sneeze'' more
often than a ``no'' for more prototypical ``brush'').

Investigating the distribution of verbs over the output
classes for GPT-4, we indeed find that the most frequent
five verbs of the Y$\rightarrow$Y are all highly prototypical motion
verbs: \textit{fling, chuck, pull, teleport,  slam}.
In contrast, the most frequent five verbs for N$\rightarrow$Y
are not: \textit{steal, eat, tie, smash, separate}.

\begin{table}
\centering
\begin{tabular}{lllrrr}
\toprule
Family& Model & IT & Y$\rightarrow$Y&  N$\rightarrow$Y& X$\rightarrow$N\\
\midrule
\multirow{2}{*}{GPT} &3.5& + &43.20 & 10.70&46.10\\
&4& + & 57.07& 11.23&31.70\\
\midrule
Gemini&Pro& + & 43.43& 12.70&43.87\\
\midrule
\multirow{6}{*}{Llama2} &\multirow{2}{*}{7b}&-- & 9.54& 1.09&89.37\\
&& +&21.93& 1.77&76.29\\
&\multirow{2}{*}{13b}& --&53.00& 8.72&38.28\\
&& +&5.59& 1.23&93.19\\
&\multirow{2}{*}{70b$_Q$}& --&36.65& 7.36&55.99\\
&&+& 37.87& 5.59&56.54\\
\midrule
\multirow{4}{*}{Mistral} &\multirow{2}{*}{7b}& -- & 34.20&4.50 &61.31\\
&& + & 68.12& 8.45&23.43\\
& \multirow{2}{*}{8x7b}&-- & 35.29& 9.95&54.77\\
&& + & 69.75& 12.13&18.12\\

\bottomrule
\end{tabular}
\caption{LLM evaluation results. IT=instruction-tuned. Q=quantised.}
\label{tab:gpt_results}
\end{table}
\section{Conclusion}

Our paper has made several contributions. We have introduced
an annotation pipeline aided by dependency parsing and
prompting GPT-3.5, which can be specifically used for
phenomena that are so rare that little to no corpora have
been created, as the human annotation effort would be
too great. We have demonstrated this pipeline on the example
of the caused-motion construction, and created corpora of 765
manually annotated
and 127,955 automatically
annotated (but high-confidence)
CMC examples.
 We have used the manually annotated corpus to evaluate
state-of-the-art LLMs for their understanding of the CMC,
and found that they have
high error rates ($>$30\%) when asked to interpret
situations
described with a non-prototypical CMC.

We hope that our work will  inspire more
computational and corpus-based studies of rare linguistic
phenomena. We note that even though
prompt engineering is complex, large gains can
be achieved by using intermediate-complexity prompts and
basic
knowledge of LLMs. We are confident that
further advances in instruction-tuned LLMs will make the
cost-benefit ratio of incorporating them into this hybrid annotation
pipeline even stronger.

We see several opportunities for interesting future work in
both halves of the paper. For the data collection part, it is a  promising engineering direction to
develop tools that automate parts of this process so that it
becomes available to linguists without
the need for complex prompt engineering. Continued progress
in LLMs is likely to
make the process
even more efficient.

Concerning the evaluation of LLMs' understanding of
constructions, a straightforward direction for future work
would be to expand to the other three Argument Structure
Constructions described in \citet{goldberg1992}.

\section*{Limitations} 
The data collection section of our work is limited because it draws on GPT-3.5, a commercial non-open model. While the general principles and guidelines for data collection hold for any LLM that is used for data annotation, some of the specific takeaways, especially the cost formulation, are specific to GPT-3.5. Further, the final size of our dataset is limited by our budget for prompting GPT-3.5, as well as the time available for annotation. 
\bibliography{anthology,custom}
\bibliographystyle{acl_natbib}

\appendix
\section{Detailed Graphs of Prompt and Total Costs}

\begin{figure*}
    \centering
    \includegraphics[width=\textwidth]{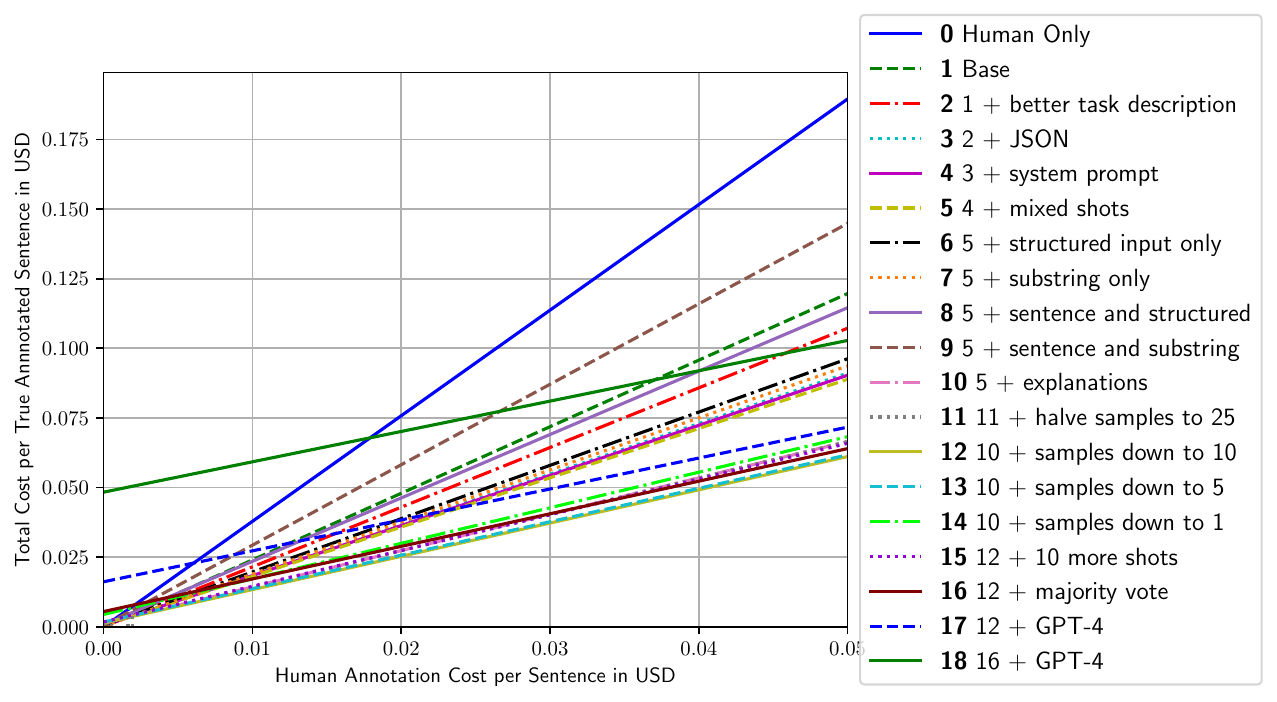}
    \caption{Total cost per final true annotated example compared to the cost of one human sentence annotation}
    \label{fig:costs_full}
\end{figure*}

\begin{figure*}
    \centering
    \includegraphics[width=\textwidth]{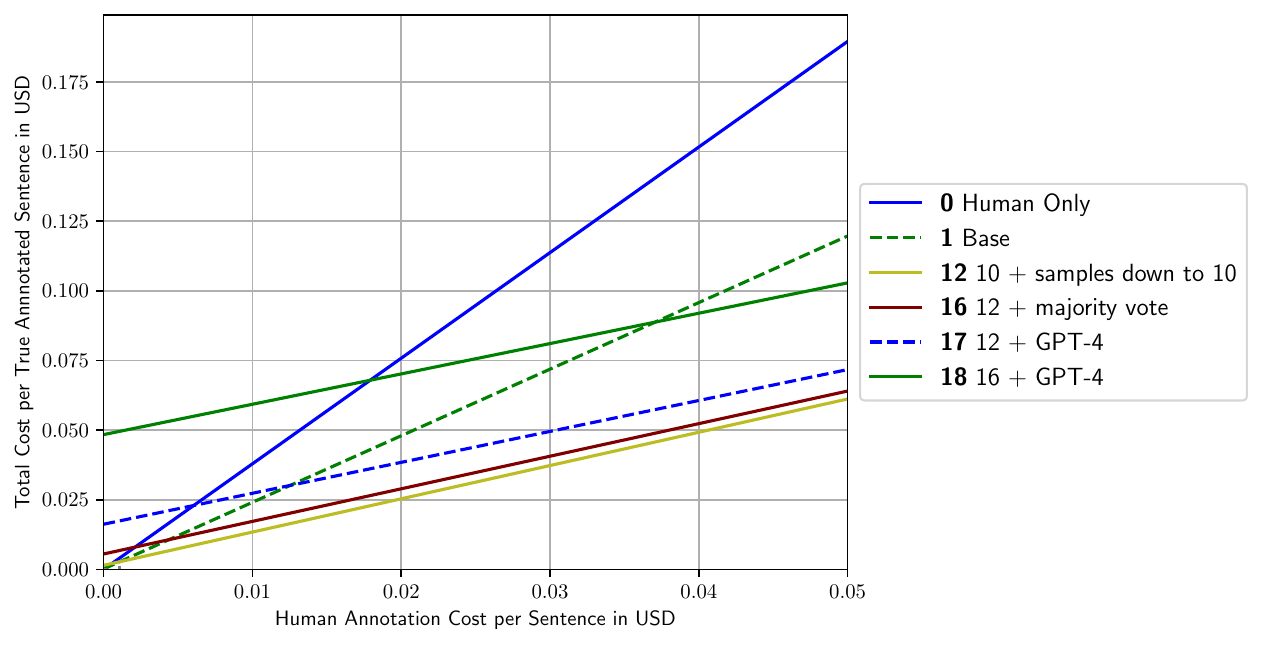}
    \caption{Total cost per final true annotated example compared to the cost of one human sentence annotation, with only prompts 0 (fully manual human baseline), 1 (base prompt), 12, 16, 17, and 18 included}
    \label{fig:costs_partial_small}
\end{figure*}

\begin{figure*}
    \centering
    \includegraphics[width=\textwidth]{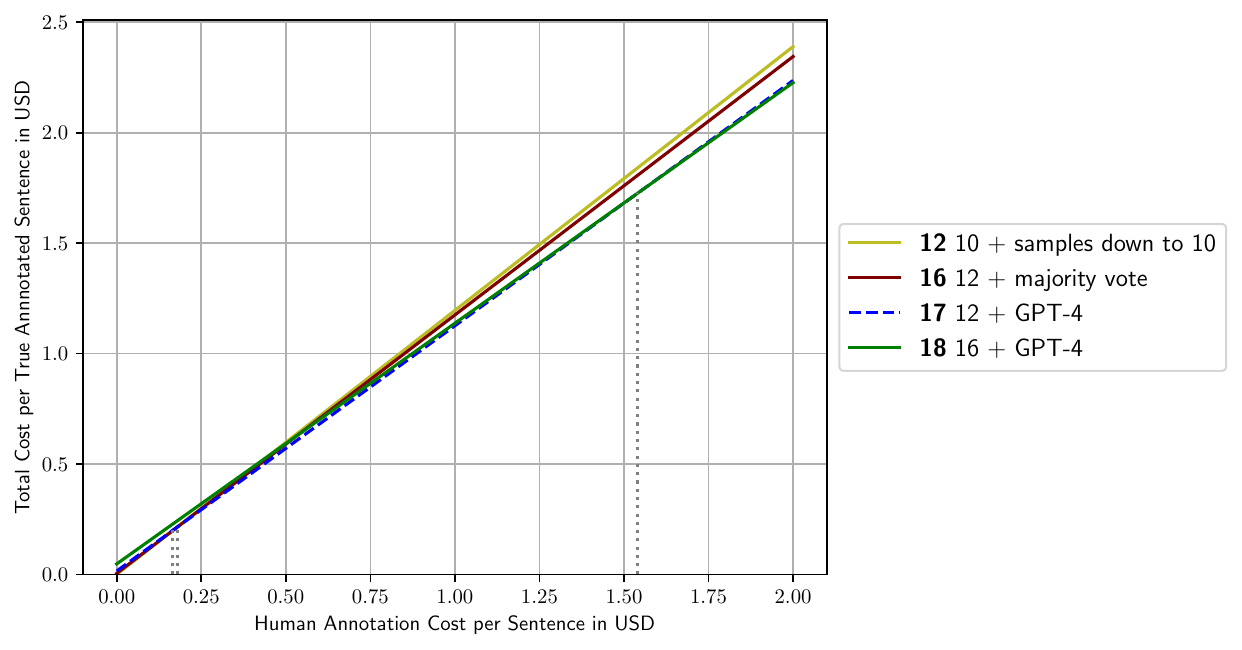}
    \caption{Total cost per final true annotated example compared to the cost of one human sentence annotation, focused on the points where the optimal prompt changes from 12 to 16, 16 to 17, and 17 to 18}
    \label{fig:costs_partial_big}
\end{figure*}

We include a visual representation of the cost of each prompt dependent on the human annotation cost per sentence. All prompts are displayed in Figure \ref{fig:costs_full}. In Figure \ref{fig:costs_partial_small}, we remove most of the worse-performing prompts and show the different slopes and intercept points. In Figure \ref{fig:costs_partial_big}, we highlight the points in the human annotation cost where the optimal prompt changes.

\section{Full Statistics on each prompt}
\label{full_matrices}

\begin{table}[h]
\small
\centering
\begin{tabular}{lrrrrr}
\toprule
\textbf{Prompt} & \textbf{Prec.} & \textbf{Rec.} & \textbf{F1}  & \textbf{HR \$} & \textbf{GPT \$ in ct} \\
\midrule
1 & 41.84 & 57.89 & 48.58 & 2.389 & 0.01 \\
2 & 46.70 & 69.69 & 55.92 & 2.141 & 0.01 \\
3 & 55.13 & 76.69 & 64.15 & 1.813 & 0.04 \\
4 & 55.61 & 78.19 & 65.00 & 1.798 & 0.04 \\
5 & 56.54 & 71.42 & 63.12 & 1.768 & 0.04 \\
6 & 52.28 & 77.44 & 62.42 & 1.912 & 0.06 \\
7 & 53.53 & 79.69 & 64.04 & 1.867 & 0.03 \\
8 & 43.93 & 87.21 & 58.43 & 2.275 & 0.07 \\
9 & 34.63 & 93.23 & 50.50 & 2.887 & 0.04 \\
10 & 76.62 & 44.36 & 56.19 & 1.305 & 0.13 \\
11 & 76.54 & 46.61 & 57.94 & 1.306 & 0.12 \\
12 & 83.75 & 50.37 & 62.91 & 1.194 & 0.14 \\
13 & 83.33 & 52.63 & 64.51 & 1.200 & 0.17 \\
14 & 78.35 & 57.14 & 66.08 & 1.276 & 0.44 \\
15 & 78.02 & 53.38 & 63.39 & 1.281 & 0.18 \\
16 & 85.50 & 44.36 & 58.41 & 1.169 & 0.55 \\
17 & 90.09 & 75.18 & 81.96 & 1.110 & 1.62\\
 18& 91.81& 75.93& 83.12& 1.089&4.83\\
\bottomrule

\end{tabular}
\caption{Full Evaluation Results for Every Prompt}
\label{tab:prompt_results_appendix}
\end{table}

In Table \ref{tab:prompt_results_appendix}, we report the sequence of prompts that we tried and their performance. Details of all prompts and the changes from the previous prompt are available in Appendix Section \ref{prompt_details}.

\section{Full details for each prompt}
\label{prompt_details}

We report in Tables \ref{tab:prompt_1} to \ref{tab:prompt_18} the details of the prompt, along with the change that it represents from a previous prompt.

\begin{table*}
\small
    \centering
    \begin{tabularx}{\textwidth}{lX}
    \toprule
         \textbf{System Prompt}& -\\
         \textbf{Instruction}&                    The task is to classify whether the sentences are instances of the caused motion construction as first introduced by Goldberg (1992) or not.\\
         \textbf{Input Format}& Here are 5 positive examples: \textit{id}, \textit{sentence}, \textit{label}
Here are 5 negative examples: \textit{id}, \textit{sentence}, \textit{label}. 
Classify the following sentences: \textit{id}, \textit{sentence}\\
         \textbf{Output Format}& Reply with a csv codeblock (wrapped in three backticks), with the headers 'id' and 'label'. label should be either True or False. Label all 50 sentences.\\
         \textbf{Shots per Class}& 5\\
         \textbf{Sentences}& 50\\
 \textbf{Model}&GPT-3.5\\
 \textbf{Majority Vote}&No\\
   \bottomrule
    \end{tabularx}
    \caption{Prompt 1}
    \label{tab:prompt_1}
\end{table*}

\begin{table*}
\small
    \centering
    \begin{tabularx}{\textwidth}{lX}
    \toprule
         \textbf{System Prompt}& -\\
         \textbf{Instruction}&                    The task is to classify whether the sentences are instances of the caused motion construction as first introduced by Goldberg (1992) or not.
A caused-motion construction is a linguistic phenomenon where a verb describes an action that results in a change of location or motion for a specific object. 
Your task will be to understand what is going on in the sentence and determine if the verb describes an action that results in a change of location or motion for a specific object.
Keep in mind that the caused-motion construction is rare, and label the sentences accordingly.\\
         \textbf{Input Format}& Here are 5 positive examples: \textit{id}, \textit{sentence}, \textit{label}
Here are 5 negative examples: \textit{id}, \textit{sentence}, \textit{label}. 
Classify the following sentences: \textit{id}, \textit{sentence}\\
         \textbf{Output Format}& Reply with a csv codeblock (wrapped in three backticks), with the headers 'id' and 'label'. label should be either True or False. Label all 50 sentences.\\
         \textbf{Shots per Class}& 5\\
         \textbf{Sentences}& 50\\
 \textbf{Model}&GPT-3.5\\
 \textbf{Majority Vote}&No\\
  \textbf{Change}&Prompt 1 with longer instruction \\
    \bottomrule
    \end{tabularx}
    \caption{Prompt 2.}
    \label{tab:prompt_2}
\end{table*}

\begin{table*}
\small
    \centering
    \begin{tabularx}{\textwidth}{lX}
    \toprule
         \textbf{System Prompt}& -\\
         \textbf{Instruction}&                    The task is to classify whether the sentences are instances of the caused motion construction as first introduced by Goldberg (1992) or not.
A caused-motion construction is a linguistic phenomenon where a verb describes an action that results in a change of location or motion for a specific object. 
Your task will be to understand what is going on in the sentence and determine if the verb describes an action that results in a change of location or motion for a specific object.
Keep in mind that the caused-motion construction is rare, and label the sentences accordingly.\\
\textbf{Input Format}& Here are 5 positive examples: \{ "id": \textit{id},"sentence": \textit{sentence}, "label": \textit{label} \}. 
Here are 5 negative examples: \{ "id": \textit{id},"sentence": \textit{sentence}, "label": \textit{label} \}. 
Classify the following sentences: \{ "id": \textit{id},"sentence": \textit{sentence} \}. \\
\textbf{Output Format}& Respond with a jsonl codeblock (wrapped in three backticks). Each object should include an "id", "sentence", and finally a "label" field with either "true" or "false". Label all 50 sentences.\\
\textbf{Shots per Class}& 5\\
\textbf{Sentences}& 50\\
 \textbf{Model}&GPT-3.5\\
 \textbf{Majority Vote}&No\\
  \textbf{Change}&Prompt 2 with input and output format changed to JSON \\
    \bottomrule
    \end{tabularx}
    \caption{Prompt 3}
    \label{tab:prompt_3}
\end{table*}

\begin{table*}
\small
    \centering
    \begin{tabularx}{\textwidth}{lX}
    \toprule
         \textbf{System Prompt}& You are a linguistic expert specializing in syntax, specifically the caused-motion construction in English sentences. Your task is to analyze given sentences and classify whether they exhibit this construction or not. Remember to carefully consider the structure and meaning of each sentence to make the most accurate determination.\\
         \textbf{Instruction}&                    The task is to classify whether the sentences are instances of the caused motion construction as first introduced by Goldberg (1992) or not.
A caused-motion construction is a linguistic phenomenon where a verb describes an action that results in a change of location or motion for a specific object. 
Your task will be to understand what is going on in the sentence and determine if the verb describes an action that results in a change of location or motion for a specific object.
Keep in mind that the caused-motion construction is rare, and label the sentences accordingly.\\
\textbf{Input Format}& Here are 5 positive examples: \{ "id": \textit{id},"sentence": \textit{sentence}, "label": \textit{label} \}. 
Here are 5 negative examples: \{ "id": \textit{id},"sentence": \textit{sentence}, "label": \textit{label} \}. 
Classify the following sentences: \{ "id": \textit{id},"sentence": \textit{sentence} \}. \\
\textbf{Output Format}& Respond with a jsonl codeblock (wrapped in three backticks). Each object should include an "id", "sentence", and finally a "label" field with either "true" or "false". Label all 50 sentences.\\
\textbf{Shots per Class}& 5\\
\textbf{Sentences}& 50\\
 \textbf{Model}&GPT-3.5\\
 \textbf{Majority Vote}&No\\
  \textbf{Change}&Prompt 3 with added system prompt \\
    \bottomrule
    \end{tabularx}
    \caption{Prompt 4}
    \label{tab:prompt_4}
\end{table*}

\begin{table*}
\small
    \centering
    \begin{tabularx}{\textwidth}{lX}
    \toprule
         \textbf{System Prompt}& You are a linguistic expert specializing in syntax, specifically the caused-motion construction in English sentences. Your task is to analyze given sentences and classify whether they exhibit this construction or not. Remember to carefully consider the structure and meaning of each sentence to make the most accurate determination.\\
         \textbf{Instruction}&                    The task is to classify whether the sentences are instances of the caused motion construction as first introduced by Goldberg (1992) or not.
A caused-motion construction is a linguistic phenomenon where a verb describes an action that results in a change of location or motion for a specific object. 
Your task will be to understand what is going on in the sentence and determine if the verb describes an action that results in a change of location or motion for a specific object.
Keep in mind that the caused-motion construction is rare, and label the sentences accordingly.\\
\textbf{Input Format}& Here are 10 examples with ground truth labels: : \{ "id": \textit{id},"sentence": \textit{sentence}, "label": \textit{label} \}.  
Classify the following sentences: \{ "id": \textit{id},"sentence": \textit{sentence} \}. \\
\textbf{Output Format}& Respond with a jsonl codeblock (wrapped in three backticks). Each object should include an "id", "sentence", and finally a "label" field with either "true" or "false". Label all 50 sentences.\\
\textbf{Shots per Class}& 5\\
\textbf{Sentences}& 50\\
 \textbf{Model}&GPT-3.5\\
 \textbf{Majority Vote}&No\\
  \textbf{Change}&Prompt 4 with few shots alternating between True and False labels. \\
    \bottomrule
    \end{tabularx}
    \caption{Prompt 5}
    \label{tab:prompt_5}
\end{table*}

\begin{table*}
\small
    \centering
    \begin{tabularx}{\textwidth}{lX}
    \toprule
         \textbf{System Prompt}& You are a linguistic expert specializing in syntax, specifically the caused-motion construction in English sentences. Your task is to analyze given sentences and classify whether they exhibit this construction or not. Remember to carefully consider the structure and meaning of each sentence to make the most accurate determination.\\
         \textbf{Instruction}&                    The task is to classify whether the sentences are instances of the caused motion construction as first introduced by Goldberg (1992) or not.
A caused-motion construction is a linguistic phenomenon where a verb describes an action that results in a change of location or motion for a specific object. 
Your task will be to understand what is going on in the sentence and determine if the verb describes an action that results in a change of location or motion for a specific object.
Keep in mind that the caused-motion construction is rare, and label the sentences accordingly.\\
\textbf{Input Format}& Here are 10 examples with ground truth labels: : \{ "id": \textit{id}, "verb" : \textit{verb}, "direct object": \textit{direct object}, "preposition" : \textit{preposition}, "prepositional object" : \textit{prepositional object}, "label": \textit{label} \}.  
Classify the following sentences: \{ "id": \textit{id}, "verb" : \textit{verb}, "direct object": \textit{direct object}, "preposition" : \textit{preposition}, "prepositional object" : \textit{prepositional object} \}. \\
\textbf{Output Format}& Respond with a jsonl codeblock (wrapped in three backticks). Each object should include an "id", "verb", "direct object", "preposition", "prepositional object", and finally a "label" field with either "true" or "false". Label all 50 sentences.\\
\textbf{Shots per Class}& 5\\
\textbf{Sentences}& 50\\
 \textbf{Model}&GPT-3.5\\
 \textbf{Majority Vote}&No\\
  \textbf{Change}&Prompt 5 with the sentences replaced by the information extracted with the dependency filtering step \\
    \bottomrule
    \end{tabularx}
    \caption{Prompt 6}
    \label{tab:prompt_6}
\end{table*}

\begin{table*}
\small
    \centering
    \begin{tabularx}{\textwidth}{lX}
    \toprule
         \textbf{System Prompt}& You are a linguistic expert specializing in syntax, specifically the caused-motion construction in English sentences. Your task is to analyze given sentences and classify whether they exhibit this construction or not. Remember to carefully consider the structure and meaning of each sentence to make the most accurate determination.\\
         \textbf{Instruction}&                    The task is to classify whether the sentences are instances of the caused motion construction as first introduced by Goldberg (1992) or not.
A caused-motion construction is a linguistic phenomenon where a verb describes an action that results in a change of location or motion for a specific object. 
Your task will be to understand what is going on in the sentence and determine if the verb describes an action that results in a change of location or motion for a specific object.
Keep in mind that the caused-motion construction is rare, and label the sentences accordingly.\\
\textbf{Input Format}& Here are 10 examples with ground truth labels: : \{ "id": \textit{id}, "string" : \textit{verb direct object preposition prepositional object}, "label": \textit{label} \}.  
Classify the following sentences: \{ "id": \textit{id}, "string" : \textit{verb direct object preposition prepositional object} \}. \\
\textbf{Output Format}& Respond with a jsonl codeblock (wrapped in three backticks). Each object should include an "id", a "string", and finally a "label" field with either "true" or "false". Label all 50 sentences.\\
\textbf{Shots per Class}& 5\\
\textbf{Sentences}& 50\\
 \textbf{Model}&GPT-3.5\\
 \textbf{Majority Vote}&No\\
  \textbf{Change}&Prompt 5 with the sentences replaced by the substring from the verb to the prepositional object \\
    \bottomrule
    \end{tabularx}
    \caption{Prompt 7}
    \label{tab:prompt_7}
\end{table*}

\begin{table*}
\small
    \centering
    \begin{tabularx}{\textwidth}{lX}
    \toprule
         \textbf{System Prompt}& You are a linguistic expert specializing in syntax, specifically the caused-motion construction in English sentences. Your task is to analyze given sentences and classify whether they exhibit this construction or not. Remember to carefully consider the structure and meaning of each sentence to make the most accurate determination.\\
         \textbf{Instruction}&                    The task is to classify whether the sentences are instances of the caused motion construction as first introduced by Goldberg (1992) or not.
A caused-motion construction is a linguistic phenomenon where a verb describes an action that results in a change of location or motion for a specific object. 
Your task will be to understand what is going on in the sentence and determine if the verb describes an action that results in a change of location or motion for a specific object.
Keep in mind that the caused-motion construction is rare, and label the sentences accordingly.\\
\textbf{Input Format}& Here are 10 examples with ground truth labels: : \{ "id": \textit{id}, "sentence" : \textit{sentence}, "verb" : \textit{verb}, "direct object": \textit{direct object}, "preposition" : \textit{preposition}, "prepositional object" : \textit{prepositional object}, "label": \textit{label} \}.  
Classify the following sentences: \{ "id": \textit{id}, "sentence" : \textit{sentence}, "verb" : \textit{verb}, "direct object": \textit{direct object}, "preposition" : \textit{preposition}, "prepositional object" : \textit{prepositional object} \}. \\
\textbf{Output Format}& Respond with a jsonl codeblock (wrapped in three backticks). Each object should include an "id", "sentence", "verb", "direct object", "preposition", "prepositional object", and finally a "label" field with either "true" or "false". Label all 50 sentences.\\
\textbf{Shots per Class}& 5\\
\textbf{Sentences}& 50\\
 \textbf{Model}&GPT-3.5\\
 \textbf{Majority Vote}&No\\
  \textbf{Change}&Prompt 6 with the full sentence added back in \\
    \bottomrule
    \end{tabularx}
    \caption{Prompt 8}
    \label{tab:prompt_8}
\end{table*}

\begin{table*}
\small
    \centering
    \begin{tabularx}{\textwidth}{lX}
    \toprule
         \textbf{System Prompt}& You are a linguistic expert specializing in syntax, specifically the caused-motion construction in English sentences. Your task is to analyze given sentences and classify whether they exhibit this construction or not. Remember to carefully consider the structure and meaning of each sentence to make the most accurate determination.\\
         \textbf{Instruction}&                    The task is to classify whether the sentences are instances of the caused motion construction as first introduced by Goldberg (1992) or not.
A caused-motion construction is a linguistic phenomenon where a verb describes an action that results in a change of location or motion for a specific object. 
Your task will be to understand what is going on in the sentence and determine if the verb describes an action that results in a change of location or motion for a specific object.
Keep in mind that the caused-motion construction is rare, and label the sentences accordingly.\\
\textbf{Input Format}& Here are 10 examples with ground truth labels: : \{ "id": \textit{id}, "sentence": \textit{sentence}, "string" : \textit{verb direct object preposition prepositional object}, "label": \textit{label} \}.  
Classify the following sentences: \{ "id": \textit{id}, "sentence": \textit{sentence},  "string" : \textit{verb direct object preposition prepositional object} \}. \\
\textbf{Output Format}& Respond with a jsonl codeblock (wrapped in three backticks). Each object should include an "id", a "sentence", a "string", and finally a "label" field with either "true" or "false". Label all 50 sentences.\\
\textbf{Shots per Class}& 5\\
\textbf{Sentences}& 50\\
 \textbf{Model}&GPT-3.5\\
 \textbf{Majority Vote}&No\\
  \textbf{Change}&Prompt 7 with the full sentence added back in \\
    \bottomrule
    \end{tabularx}
    \caption{Prompt 9}
    \label{tab:prompt_9}
\end{table*}

\begin{table*}
\small
    \centering
    \begin{tabularx}{\textwidth}{lX}
    \toprule
         \textbf{System Prompt}& You are a linguistic expert specializing in syntax, specifically the caused-motion construction in English sentences. Your task is to analyze given sentences and classify whether they exhibit this construction or not. Remember to carefully consider the structure and meaning of each sentence to make the most accurate determination.\\
         \textbf{Instruction}&                    The task is to classify whether the sentences are instances of the caused motion construction as first introduced by Goldberg (1992) or not.
A caused-motion construction is a linguistic phenomenon where a verb describes an action that results in a change of location or motion for a specific object. 
Your task will be to understand what is going on in the sentence and determine if the verb describes an action that results in a change of location or motion for a specific object.
Keep in mind that the caused-motion construction is rare, and label the sentences accordingly.\\
\textbf{Input Format}& Here are 10 examples with examples with explanations and ground truth labels: : \{ "id": \textit{id},"sentence": \textit{sentence}, "explanation": \textit{explanation}, "label": \textit{label} \}.  
Classify the following sentences: \{ "id": \textit{id},"sentence": \textit{sentence} \}. \\
\textbf{Output Format}& Respond with a jsonl codeblock (wrapped in three backticks). Each object should include an "id", "sentence", "explanation", and finally a "label" field with either "true" or "false". Label all 50 sentences.\\
\textbf{Shots per Class}& 5\\
\textbf{Sentences}& 50\\
 \textbf{Model}&GPT-3.5\\
 \textbf{Majority Vote}&No\\
  \textbf{Change}&Prompt 5 with explanations for the labels added to input and required from the model for the output \\
    \bottomrule
    \end{tabularx}
    \caption{Prompt 10}
    \label{tab:prompt_10}
\end{table*}

\begin{table*}
\small
    \centering
    \begin{tabularx}{\textwidth}{lX}
    \toprule
         \textbf{System Prompt}& You are a linguistic expert specializing in syntax, specifically the caused-motion construction in English sentences. Your task is to analyze given sentences and classify whether they exhibit this construction or not. Remember to carefully consider the structure and meaning of each sentence to make the most accurate determination.\\
         \textbf{Instruction}&                    The task is to classify whether the sentences are instances of the caused motion construction as first introduced by Goldberg (1992) or not.
A caused-motion construction is a linguistic phenomenon where a verb describes an action that results in a change of location or motion for a specific object. 
Your task will be to understand what is going on in the sentence and determine if the verb describes an action that results in a change of location or motion for a specific object.
Keep in mind that the caused-motion construction is rare, and label the sentences accordingly.\\
\textbf{Input Format}& Here are 10 examples with examples with explanations and ground truth labels: : \{ "id": \textit{id},"sentence": \textit{sentence}, "explanation": \textit{explanation}, "label": \textit{label} \}.  
Classify the following sentences: \{ "id": \textit{id},"sentence": \textit{sentence} \}. \\
\textbf{Output Format}& Respond with a jsonl codeblock (wrapped in three backticks). Each object should include an "id", "sentence", "explanation", and finally a "label" field with either "true" or "false". Label all 50 sentences.\\
\textbf{Shots per Class}& 5\\
\textbf{Sentences}& 50 $\rightarrow$ 25 $\rightarrow$ 10 $\rightarrow$ 5 $\rightarrow$ 1\\
 \textbf{Model}&GPT-3.5\\
 \textbf{Majority Vote}&No\\
  \textbf{Change}&Prompt 10 with different numbers of sentences classified per prompt \\
    \bottomrule
    \end{tabularx}
    \caption{Prompts 11-14}
    \label{tab:prompt_11_14}
\end{table*}

\begin{table*}
\small
    \centering
    \begin{tabularx}{\textwidth}{lX}
    \toprule
         \textbf{System Prompt}& You are a linguistic expert specializing in syntax, specifically the caused-motion construction in English sentences. Your task is to analyze given sentences and classify whether they exhibit this construction or not. Remember to carefully consider the structure and meaning of each sentence to make the most accurate determination.\\
         \textbf{Instruction}&                    The task is to classify whether the sentences are instances of the caused motion construction as first introduced by Goldberg (1992) or not.
A caused-motion construction is a linguistic phenomenon where a verb describes an action that results in a change of location or motion for a specific object. 
Your task will be to understand what is going on in the sentence and determine if the verb describes an action that results in a change of location or motion for a specific object.
Keep in mind that the caused-motion construction is rare, and label the sentences accordingly.\\
\textbf{Input Format}& Here are 20 examples with examples with explanations and ground truth labels: : \{ "id": \textit{id},"sentence": \textit{sentence}, "explanation": \textit{explanation}, "label": \textit{label} \}.  
Classify the following sentences: \{ "id": \textit{id},"sentence": \textit{sentence} \}. \\
\textbf{Output Format}& Respond with a jsonl codeblock (wrapped in three backticks). Each object should include an "id", "sentence", "explanation", and finally a "label" field with either "true" or "false". Label all 50 sentences.\\
\textbf{Shots per Class}& 10\\
\textbf{Sentences}& 10\\
 \textbf{Model}&GPT-3.5\\
 \textbf{Majority Vote}&No\\
  \textbf{Change}&Prompt 12 with the number of shots per class doubled from 5 to 10 \\
    \bottomrule
    \end{tabularx}
    \caption{Prompt 15}
    \label{tab:prompt_15}
\end{table*}

\begin{table*}
\small
    \centering
    \begin{tabularx}{\textwidth}{lX}
    \toprule
         \textbf{System Prompt}& You are a linguistic expert specializing in syntax, specifically the caused-motion construction in English sentences. Your task is to analyze given sentences and classify whether they exhibit this construction or not. Remember to carefully consider the structure and meaning of each sentence to make the most accurate determination.\\
         \textbf{Instruction}&                    The task is to classify whether the sentences are instances of the caused motion construction as first introduced by Goldberg (1992) or not.
A caused-motion construction is a linguistic phenomenon where a verb describes an action that results in a change of location or motion for a specific object. 
Your task will be to understand what is going on in the sentence and determine if the verb describes an action that results in a change of location or motion for a specific object.
Keep in mind that the caused-motion construction is rare, and label the sentences accordingly.\\
\textbf{Input Format}& Here are 10 examples with examples with explanations and ground truth labels: : \{ "id": \textit{id},"sentence": \textit{sentence}, "explanation": \textit{explanation}, "label": \textit{label} \}.  
Classify the following sentences: \{ "id": \textit{id},"sentence": \textit{sentence} \}. \\
\textbf{Output Format}& Respond with a jsonl codeblock (wrapped in three backticks). Each object should include an "id", "sentence", "explanation", and finally a "label" field with either "true" or "false". Label all 50 sentences.\\
\textbf{Shots per Class}& 5\\
\textbf{Sentences}& 10\\
 \textbf{Model}&GPT-3.5\\
 \textbf{Majority Vote}&Yes\\
  \textbf{Change}&Prompt 12 with a majority vote from three separate runs \\
    \bottomrule
    \end{tabularx}
    \caption{Prompt 16}
    \label{tab:prompt_16}
\end{table*}

\begin{table*}
\small
    \centering
    \begin{tabularx}{\textwidth}{lX}
    \toprule
         \textbf{System Prompt}& You are a linguistic expert specializing in syntax, specifically the caused-motion construction in English sentences. Your task is to analyze given sentences and classify whether they exhibit this construction or not. Remember to carefully consider the structure and meaning of each sentence to make the most accurate determination.\\
         \textbf{Instruction}&                    The task is to classify whether the sentences are instances of the caused motion construction as first introduced by Goldberg (1992) or not.
A caused-motion construction is a linguistic phenomenon where a verb describes an action that results in a change of location or motion for a specific object. 
Your task will be to understand what is going on in the sentence and determine if the verb describes an action that results in a change of location or motion for a specific object.
Keep in mind that the caused-motion construction is rare, and label the sentences accordingly.\\
\textbf{Input Format}& Here are 10 examples with examples with explanations and ground truth labels: : \{ "id": \textit{id},"sentence": \textit{sentence}, "explanation": \textit{explanation}, "label": \textit{label} \}.  
Classify the following sentences: \{ "id": \textit{id},"sentence": \textit{sentence} \}. \\
\textbf{Output Format}& Respond with a jsonl codeblock (wrapped in three backticks). Each object should include an "id", "sentence", "explanation", and finally a "label" field with either "true" or "false". Label all 50 sentences.\\
\textbf{Shots per Class}& 5\\
\textbf{Sentences}& 10\\
 \textbf{Model}&GPT-4\\
 \textbf{Majority Vote}&No\\
  \textbf{Change}&Prompt 12 using GPT-4 \\
    \bottomrule
    \end{tabularx}
    \caption{Prompt 17}
    \label{tab:prompt_17}
\end{table*}

\begin{table*}
\small
    \centering
    \begin{tabularx}{\textwidth}{lX}
    \toprule
         \textbf{System Prompt}& You are a linguistic expert specializing in syntax, specifically the caused-motion construction in English sentences. Your task is to analyze given sentences and classify whether they exhibit this construction or not. Remember to carefully consider the structure and meaning of each sentence to make the most accurate determination.\\
         \textbf{Instruction}&                    The task is to classify whether the sentences are instances of the caused motion construction as first introduced by Goldberg (1992) or not.
A caused-motion construction is a linguistic phenomenon where a verb describes an action that results in a change of location or motion for a specific object. 
Your task will be to understand what is going on in the sentence and determine if the verb describes an action that results in a change of location or motion for a specific object.
Keep in mind that the caused-motion construction is rare, and label the sentences accordingly.\\
\textbf{Input Format}& Here are 10 examples with examples with explanations and ground truth labels: : \{ "id": \textit{id},"sentence": \textit{sentence}, "explanation": \textit{explanation}, "label": \textit{label} \}.  
Classify the following sentences: \{ "id": \textit{id},"sentence": \textit{sentence} \}. \\
\textbf{Output Format}& Respond with a jsonl codeblock (wrapped in three backticks). Each object should include an "id", "sentence", "explanation", and finally a "label" field with either "true" or "false". Label all 50 sentences.\\
\textbf{Shots per Class}& 5\\
\textbf{Sentences}& 10\\
 \textbf{Model}&GPT-4\\
 \textbf{Majority Vote}&Yes\\
  \textbf{Change}&Prompt 16 using GPT-4 \\
  \bottomrule
    \end{tabularx}
    \caption{Prompt 18}
    \label{tab:prompt_18}
\end{table*}

\section{Few Shots}

In Table \ref{tab:few_shots}, we give the five shots from each class given to ChatGPT as examples.

\begin{table*}
    \centering
    \small
    \begin{tabularx}{\textwidth}{XlllllX}
    \toprule
         Sentence&  Verb&  Dir Obj&  Prep&  P-Obj&  Lab.&  Explanation\\
         \midrule
         Sam sneezed the napkin off the table .&  sneeze&  napkin&  off&  table&  True&  This is caused-motion, because Sam sneezing is causing the napkin to move off the table.\\
         Joey grated the cheese onto a serving plate .&  grate&  cheese&  onto&  plate&  True&  This is caused-motion, because the grating is causing the cheese to move onto the plate.\\
         Sam assisted her out of the room .&  assist&  she&  out of&  room&  True&  This is caused-motion, because Sam assisting is causing her to move out of the room.\\
         He nudged the golf ball into the hole .&  nudge&  ball&  into&  hole&  True&  This is caused-motion, because him nudging the ball is causing it to move into the hole.\\
         Frank squeezed the ball through the crack .&  squeeze&  ball&  through&  crack&  True&  This is caused-motion, because Frank is moving the ball through the whole by squeezing it.\\
         The hammer broke the vase into pieces.&  break&  vase&  into&  piece&  False&  This is not caused-motion, because the vase is changing its state into pieces, the pieces are not a destination.\\
         Christy blew Sam under the table .&  instruct&  he&  into&  room&  False&  This is not caused-motion, because you are not moving under the table because Christy is blowing, the blowing action is taking place under the table.\\
         Adele raised her eyebrows at Sam .&  raise&  eyebrow&  at&  I&  False&  This is not caused-motion, because while Adele is moving her eyebrows, they are not literally moving towards Sam.\\
         They separated people into groups .&  separate&  people&  into&  group&  False&  This is not caused-motion, because the people aren't moving towards groups, they are becoming the groups.\\
 His cane helped him into the car .& help& he& into& car& False&This is not caused-motion because the cane isn't causing the motion, it is being used as a tool to assist with the motion.\\
 \bottomrule
    \end{tabularx}
    \caption{Few Shots used in all prompts with the structured information and explanations that are used in some prompts. P-Obj stands for Prepositional Object, Dir Obj for Direct Object.}
    \label{tab:few_shots}
\end{table*}

\end{document}